# Deep Learning for Temporal Data Representation in Electronic Health Records: A Systematic Review of Challenges and Methodologies

Feng Xie[1*], Han Yuan[2*], Yilin Ning[2], Marcus Eng Hock Ong[1,2,3], Mengling Feng[4], Wynne Hsu[5,6], Bibhas Chakraborty[1,2,7,8], Nan Liu[1,2,6,9#]

[1] Programme in Health Services and Systems Research, Duke-NUS Medical School, Singapore, Singapore
[2] Centre for Quantitative Medicine, Duke-NUS Medical School, Singapore, Singapore
[3] Department of Emergency Medicine, Singapore General Hospital, Singapore, Singapore
[4] Saw Swee Hock School of Public Health, National University of Singapore, Singapore, Singapore
[5] School of Computing, National University of Singapore, Singapore, Singapore
[6] Institute of Data Science, National University of Singapore, Singapore, Singapore
[7] Department of Statistics and Data Science, National University of Singapore, Singapore, Singapore
[8] Department of Biostatistics and Bioinformatics, Duke University, Durham, NC, United States
[9] Health Services Research Centre, Singapore Health Services, Singapore, Singapore

* These authors contributed equally

# Corresponding Author:
Nan Liu
Programme in Health Services and Systems Research
Duke-NUS Medical School
8 College Road
Singapore 169857
Singapore
Phone: +65 6601 6503
Email: liu.nan@duke-nus.edu.sg

## Abstract:

**Objective**: Temporal electronic health records (EHRs) can be a wealth of information for secondary uses, such as clinical events prediction or chronic disease management. However, challenges exist for temporal data representation. We therefore sought to identify these challenges and evaluate novel methodologies for addressing them through a systematic examination of deep learning solutions.

**Methods**: We searched five databases (PubMed, EMBASE, the Institute of Electrical and Electronics Engineers [IEEE] Xplore Digital Library, the Association for Computing Machinery [ACM] digital library, and Web of Science) complemented with hand-searching in several prestigious computer science conference proceedings. We sought articles that reported deep learning methodologies on temporal data representation in structured EHR data from January 1, 2010, to August 30, 2020. We summarized and analyzed the selected articles from three perspectives: nature of time series, methodology, and model implementation.

**Results**: We included 98 articles related to temporal data representation using deep learning. Four major challenges were identified, including data irregularity, data heterogeneity, data sparsity, and model opacity. We then studied how deep learning techniques were applied to address these challenges. Finally, we discuss some open challenges arising from deep learning.

**Conclusion**: Temporal EHR data present several major challenges for clinical prediction modeling and data utilization. To some extent, current deep learning solutions can address these challenges. Future studies can consider designing comprehensive and integrated solutions. Moreover, researchers should incorporate additional clinical domain knowledge into study designs and enhance the interpretability of the model to facilitate its implementation in clinical practice.

# 1. Introduction

An electronic health record (EHR) [1] collects patients' health information in structured and unstructured digital formats. While the primary objective of an EHR is to improve the efficiency of healthcare systems, it also contains valuable information for secondary uses purposes [2]. EHR contains two types of data: structured data such as diagnoses, procedures, medication prescriptions, vital signs, lab tests, and unstructured data such as clinical notes, physiological signals, and medical images. Most structured EHR data are documented with timestamps by tracking repeated measurements of a patient's conditions over time. Compared to static data, temporal data provide longitudinal information on a patient's medical history, where hidden patterns (e.g., disease progression or changing variables over time) could be exploited. The growing amount of temporal EHR data presents an opportunity to develop more comprehensive and usable models for risk stratification, disease prognosis, or chronic disease management such as chronic kidney disease prediction [3] and adverse drug event detection [4].

Although researchers have demonstrated that incorporating temporal EHR data into predictive models can improve discriminative performance [3, 5, 6], such information is not often fully utilized due to its temporal nature [7]. Most conventional regression and machine learning methods are unable to efficiently extract the temporal pattern from data that contains multiple sets of repeated variables. Some traditional approaches rely on extracting a single value aggregated from the time series, such as mean, median, or other aggregated statistics [8]. It resulted in the loss of potentially valuable sequential information due to the inability to exploit the temporal dynamics of the data [9]. Therefore, how to better account for the temporality of time series clinical data becomes an important research question.

Temporal EHR data with complex structure and unevenly distributed clinical events present multiple technical challenges, including data irregularity, heterogeneity, sparsity, and model opacity, among others. In view of the limitations of standard learning algorithms in dealing with these challenges, the state-of-the-art deep learning-based methods, such as recurrent neural networks (RNNs) [10, 11], long short-term memory (LSTM) [12-14], and gated recurrent unit (GRU) [15], have been proposed for temporal EHR data representation. These sequential deep learning architectures present potential suitability for dealing with the temporal nature of the EHR. With their ability of learning, flexibility, and generalizability by complex nonlinearity, deep learning algorithms have demonstrated superiority when modeling temporal EHR data in many applications [16-19].

Several recent reviews have summarized the use of deep learning for analyzing general EHR data [20-23]. Nevertheless, none of them provide a systematic and in-depth summary of the technical challenges and deep learning solutions for handling temporal EHR data. In this review, we sought to consolidate the recent development of novel deep learning methods for representing temporal data and evaluate selected studies from the perspective of primary challenges and the methodologies that address them. We systematically explored the primary issues involved in analyzing temporal EHR data and thoroughly investigated state-of-the-art deep learning solutions. Moreover, we identified that there are still open challenges such as usability and transferability, which suggest potential topics for further research.

## 2. Methods

### 2.1 Search strategy and data sources

We performed a systematic review of methodological studies on the use of deep learning techniques for temporally structured EHR representations. We conducted the literature search in five databases: PubMed, EMBASE, the Institute of Electrical and Electronics Engineers (IEEE) Xplore Digital Library, the Association for Computing Machinery (ACM) Digital Library, and Web of Science. We also searched relevant articles in several prestigious computer science conference proceedings (i.e., Conference on Neural Information Processing Systems [NeurIPS], International Joint Conference on Artificial Intelligence [IJCAI], and Association for the Advancement of Artificial Intelligence [AAAI] Conference) that are not included in the above databases. The searched terms on EHR were ('electronic health record' OR 'EHR' OR 'EHRs' OR 'electronic medical record' OR 'EMR' OR 'EMRs') [7]. We also added terms ('deep learning' OR 'neural network' OR 'deep' OR 'CNN' OR 'RNN' OR 'LSTM') to limit the search to deep learning-based studies, and ('embed' OR 'embedding' OR 'representation' OR 'time series' OR 'sparse' OR 'temporal' OR 'concept' OR 'sequential' OR 'attention') to include studies of temporal data representation. We further restricted our search to papers published between January 1, 2010, and August 30, 2020. We anticipated that only few relevant articles would be published before 2010, since deep learning for EHR is a relatively new development in the last decade.

## 2.2 Inclusion and exclusion criteria

We followed the PRISMA [24] guidelines to report our systematic review. We included all methodological papers published in English, which employed deep learning for handling temporal EHR data. Review articles, duplicate records, and studies not relevant to EHR or deep learning were excluded. We further excluded pure application papers that did not propose novel methods to address challenges and papers that only dealt with static data (i.e., non-temporal data) or unstructured data (for example, free texts, physiological signals, and medical images). Two reviewers (FX and HY) independently screened all studies and, if ambiguous, discussed with NL to reach a consensus on paper selection.

## 2.3 Data extraction, synthesis, and analysis

First, we categorized included papers according to the technical challenges that they attempted to address. We identified four main categories of technical challenges in temporal EHR data analysis: data irregularity, data sparsity, data heterogeneity, and model opacity. Second, we evaluated these papers in detail from three aspects: nature of time series, methodology, and model implementation. With regard to the nature of time series, we extracted information including the time series components and the method of sparse code representation. In terms of methodology, we extracted the name of the method, the technical challenges it addressed, and the architecture of deep neural networks. For model implementation, we collected information on the clinical application, EHR datasets used (e.g., Medical Information Mart for Intensive Care [MIMIC] [25]), the main evaluation metrics (e.g., area under the receiver operating characteristic curve [AUROC]), and their main comparators. Finally, we consolidated all extracted information for subsequent analysis and investigation.

## 2.4 Definition of temporal EHR data

In this review, we present a unified definition of temporal EHR data, denoted as $D$,

$$\boldsymbol{D_i} = [v_{i,1}, v_{i,2}, v_{i,3}, ..., v_{i,T_i-1}, v_{i,T_i}] \qquad (1)$$

where $i$ represents the $i$-th patient and $T_i$ is the total number of time steps of the $i$-th patient. Different patients might have different numbers of time steps. For each time step $t$, $\boldsymbol{v_{i,t}} = <\boldsymbol{c_{i,t}}, \boldsymbol{d_{i,t}}>$, where $\boldsymbol{c_{i,t}}$ denotes the data type of sparse medical concepts (e.g., medications, diagnoses, and procedure codes) and $\boldsymbol{d}_{i,t}$ represents dense data such as lab test results and vital signs.

## 3. Results

### 3.1 Selection process and results overview

Our initial search yielded 1421 papers, of which 495 duplicates had been removed, while 926 records went through title and abstract screening. Then, 780 records were excluded as they were either not relevant to EHR (n=246), nor did they utilize deep learning methods (n=185), nor did they involve temporal data representation (n=61), just applications using existing methods (n=23), or review articles (n=22), or were based on unstructured data (n=243). As a result, we included 146 articles for full-text review. A total of 98 papers were eventually included, as shown in Table 1. Figure 1 illustrates the PRISMA diagram on literature selection.

Figure 2 summarizes the statistics for the included papers. Between 2010 and 2020, the volume of articles has increased significantly. Besides the publication date, we also identified the primary research journals and conferences. ACM Conference on Knowledge Discovery and Data Mining (n=9), IEEE International Conference on Bioinformatics and Biomedicine (n=6), IEEE International Conference on Healthcare Informatics (n=6), Journal of Biomedical Informatics (n=6) were the top four publication venues. Among the included articles, data irregularity (n=37) was the most frequently studied challenge, diagnosis (n=61) was the most commonly used temporal variable, and LSTM (n=35) was the most widely adopted deep learning architecture. Of the 98 studies, the majority (n=88) used encounters (e.g., episodes, visits, admissions) as the time step $\boldsymbol{v}_{i,t}$ , while others chose fixed time windows (e.g., one hour, day, or month) as the $\boldsymbol{v}_{i,t}$ for composing time series.

Table 1: Characteristics of the included papers in this review

| Year | Paper | Challenge | | | | Deep learning solution | | Clinical Application |
|------|-------|-----------|--|--|--|------------------------|--|----------------------|
| | | Irregularity | Sparsity | Heterogeneity | Interpretability | Method name | Main Architecture | |
| 2013 | Lasko et al.[26] | ✔ | ✔ | | | | Autoencoder | Phenotyping |
| 2015 | Esteban et al.[27] | | ✔ | | | | ANN | Clinical events prediction |
| 2015 | Mehrabi et al.[28] | ✔ | | | | | RBM | Diagnosis association discovery |
| 2015 | Tran et al.[29] | ✔ | ✔ | ✔ | | eNRBM | RBM | Suicide risk stratification |
| 2016 | Choi et al.[30] | ✔ | | | | Doctor AI | RNN | Diagnosis and medication prediction |
| 2016 | Miotto et al.[31] | | ✔ | | | Deep Patient | Autoencoder | Multiple diseases prediction |
| 2016 | Zhu et al.[32] | | | ✔ | | | CNN, Word2vec | Phenotyping |
| 2016 | Choi et al.[33] | | | | ✔ | RETAIN | RNN | Heart failure prediction |
| 2017 | Baytas et al.[12] | ✔ | | | | T-LSTM | LSTM, Autoencoder | Parkinson's disease progression prediction |
| 2017 | Che et al.[34] | | | | | ehrGAN | CNN | Heart failure and diabetes classification and data generation |
| 2017 | Che et al.[35] | | | | ✔ | GRU-D | GRU | Multiple clinical tasks |
| 2017 | Feng et al.[36] | | ✔ | | | MG-CNN | CNN | Costs and length of stay prediction |
| 2017 | Mei et al.[37] | ✔ | ✔ | | | Deep Diabetologist | RNN | Personalized hypoglycemia medication prediction |
| 2017 | Nguyen et al.[38] | ✔ | ✔ | | | Deepr | CNN | Unplanned readmission prediction |
| 2017 | Pham et al.[39] | ✔ | | | | DeepCare | LSTM | Diagnoses prediction and Intervention recommendation |

| Year | Author | | | | | Model | Architecture | Application |
|------|--------|---|---|---|---|-------|--------------|-------------|
| 2017 | Sha et al.[40] | | | | ✔ | GRNN-HA | GRU | Mortality prediction |
| 2017 | Stojanovic et al.[41] | | ✔ | | | disease+procedure2vec | Skip-gram | Healthcare quality prediction |
| 2017 | Suo et al.[42] | | | ✔ | ✔ | | GRU | Diagnosis prediction |
| 2017 | Suo et al.[43] | | | ✔ | | | CNN | Multiple disease prediction |
| 2017 | Zheng et al.[44] | ✔ | | | | | GRU | Severity scores prediction |
| 2017 | Yang et al.[18] | ✔ | | ✔ | | | LSTM, GRU | Therapy decisions prediction |
| 2017 | Yang et al.[45] | | | | ✔ | TaGiTeD | Tensor Optimization | Hospitalization and medical expense prediction |
| 2018 | Bai et al.[46] | ✔ | | | ✔ | Timeline | RNN | Disease progression prediction |
| 2018 | Cheung et al.[47] | | | ✔ | ✔ | AXCNN | CNN | Readmission prediction |
| 2018 | Le et al.[48] | ✔ | | | | DMNC | LSTM | Disease progression and drug prescription prediction |
| 2018 | Lee et al.[49] | | | | ✔ | MCA-RNN | RNN | Diagnosis prediction |
| 2018 | Lei et al.[50] | | | ✔ | | RNN-DAE | RNN, Autoencoder | Mortality prediction |
| 2018 | Lin et al.[51] | ✔ | | | | | CNN, LSTM | Sepsis prediction |
| 2018 | Ma et al.[52] | | | | ✔ | KAME | GRU | Diagnosis prediction |
| 2018 | Nguyen et al. | ✔ | | | | Resset | RNN | Diabetes and mental health prediction |
| 2018 | Park et al.[53] | | | | ✔ | FA-Attn-LSTM | LSTM | Cardiovascular disease risk prediction |
| 2018 | Park et al.[54] | | | | ✔ | COAM | RNN | Medical code prediction |
| 2018 | Rajkomar et al.[55] | | | ✔ | | | LSTM, TANN, ANN | Mortality, readmission, length of stay, diagnoses prediction |
| 2018 | Suo et al.[56] | | | ✔ | | | CNN | Phenotyping |
| 2018 | Suresh et | | | ✔ | | | LSTM, Autoencoder | Mortality prediction |

| Year | Study | | | | | Method | Architecture | Application |
|---|---|---|---|---|---|---|---|---|
| | al.[57] | | | | | | | |
| 2018 | Wu et al.[58] | | | ✔ | | | LSTM | Asthma phenotyping |
| 2018 | Xiao et al.[59] | | | | ✔ | TopicRNN | GRU | Readmissions prediction |
| 2018 | Yang et al.[60] | ✔ | | | | TGBA-F | LSTM | Septic shock prediction |
| 2018 | Zhang et al.[61] | | | | ✔ | Patient2Vec | GRU | Hospitalizations prediction |
| 2018 | Huang et al.[62] | | | ✔ | | SDAE | Autoencoder | Acute coronary syndrome risk prediction |
| 2018 | Choi et al.[63] | | ✔ | | | MiME | GRU | Multiple disease prediction |
| 2019 | An et al.[64] | | | | ✔ | DeepRisk | LSTM | Cardiovascular diseases prediction |
| 2019 | Ashfaq et al.[65] | | | ✔ | | | LSTM | Readmission prediction |
| 2019 | Fiorini et al.[66] | ✔ | | | | Tangle | LSTM | Diabetes therapy initiation prediction |
| 2019 | Guo et al.[67] | | | | ✔ | COAM | RNN | Multiple disease prediction |
| 2019 | Jun et al.[68] | ✔ | | | | | Autoencoder | Mortality prediction |
| 2019 | Kwon et al.[69] | | | | ✔ | Retain-Vis | RNN | Risk prediction model visualization |
| 2019 | Lee et al.[70] | ✔ | | | | Recent context-aware LSTM | LSTM | Clinical events prediction |
| 2019 | Li et al.[71] | ✔ | | | | VS-GRU | GRU | Mortality and disease prediction |
| 2019 | Lin et al.[72] | ✔ | | | | | LSTM | Unplanned ICU readmission prediction |
| 2019 | Liu et al.[73] | ✔ | | | | | GRU | Mortality and ICU admission prediction |
| 2019 | Liu et al.[74] | ✔ | | ✔ | | | LSTM | Sepsis prediction |
| 2019 | Macias et | ✔ | | | | | LSTM | Sepsis prediction |

| Year | Author | | | | | Model | Architecture | Task |
|---|---|---|---|---|---|---|---|---|
| | al.[75] | | | | | | | |
| 2019 | Peng et al.[76] | | ✔ | | | TeSAN | GRU | Mortality prediction |
| 2019 | Ruan et al.[77] | ✔ | | | | RNN-DAE | GRU | Mortality, comorbidity prediction and phenotyping |
| 2019 | Wang et al.[78] | ✔ | | | ✔ | MCPL-based FT-LSTM | LSTM | Clinical events prediction |
| 2019 | Wang et al.[79] | | ✔ | | | CompNet | CNN, GCN | Medication prediction |
| 2019 | Wang et al.[80] | | ✔ | | | Patient2vec | RNN | Diagnosis prediction |
| 2019 | Wang et al.[81] | ✔ | | | | MRM | LSTM | Mortality and potassium ion concentration abnormality prediction |
| 2019 | Xiang et al.[82] | | ✔ | | | | LSTM | Concept similarity analysis and disease prediction |
| 2019 | Xu et al.[83] | | | | ✔ | | RNN | Adverse cardiovascular events prediction |
| 2019 | Yang et al.[84] | | | | ✔ | GcGAN | GAN | Data generation evaluated by treatment recommendation |
| 2019 | Zhang et al.[85] | ✔ | | | | | LSTM, Autoencoder | Missing data Imputation |
| 2019 | Zhang et al.[86] | | | | ✔ | KNOWRISK | LSTM | Heart failure prediction |
| 2019 | Zhang et al.[87] | ✔ | | | | | LSTM | Septic shock prediction |
| 2019 | Zhang et al.[88] | | | | | MetaPred | CNN, LSTM | MCI, Alzheimer, Parkinson's disease prediction |
| 2020 | Afshar et al.[89] | | | ✔ | | TASTE | Tensor Factorization | Heart failure phenotyping |
| 2020 | An et al.[90] | | | ✔ | | RAHM | LSTM | Medication stocking |

| Year | Author | | | | | Method | Model | Task |
|------|--------|---|---|---|---|--------|-------|------|
| | | | | | | | | prediction |
| 2020 | Barbieri et al.[91] | ✔ | | | | | RNN | ICU readmission prediction |
| 2020 | Chu et al.[92] | | | ✔ | | DAL-EP and MTL-EP | RNN | Heart failure prediction |
| 2020 | Duan et al.[93] | ✔ | | | | | RNN | Clinical events prediction |
| 2020 | Gao et al.[94] | | | ✔ | | COMPOSE | CNN | Patient-trial matching |
| 2020 | Gao et al.[95] | | | ✔ | | StageNet | LSTM | Mortality prediction |
| 2020 | Jin et al.[96] | | | | | CarePre | RNN | Diagnosis prediction |
| 2020 | Jun et al.[97] | ✔ | | | | | RNN | Mortality prediction |
| 2020 | Landi et al.[98] | | | ✔ | | ConvAE | CNN, Autoencoder | Disease prediction |
| 2020 | Lauritsen et al.[99] | ✔ | | | | | CNN, LSTM | Sepsis prediction |
| 2020 | Li et al.[100] | | ✔ | | | GNDP | CNN | Diagnosis prediction |
| 2020 | Li et al.[101] | ✔ | | | ✔ | BEHRT | Transformer | Disease prediction |
| 2020 | Li et al.[102] | | ✔ | | | CCAE | RNN | Clinical endpoint prediction |
| 2020 | Liu et al.[103] | ✔ | ✔ | | | Medi-Care AI | RNN | Medication prediction |
| 2020 | Liu et al.[104] | | | ✔ | | RGNN | LSTM, GNN | Prescription prediction |
| 2020 | Luo et al.[105] | ✔ | | | ✔ | HiTANet | Transformer | Disease prediction |
| 2020 | Panigutti et al.[106] | | | | ✔ | DoctorXAI | RNN | Next visit prediction |
| 2020 | Qiao et al.[107] | | | ✔ | | MHM | GRU | Diagnosis prediction |
| 2020 | Rongali et al.[108] | | | ✔ | | CLOUT | LSTM | Mortality prediction |
| 2020 | Song et al.[109] | | ✔ | | | LGMNN | LSTM | Medication prediction |

| Year | Author | | | | | Name | Architecture | Task |
|------|--------|---|---|---|---|------|--------------|------|
| 2020 | Su et al.[110] | ✔ | | | | GATE | GRU | Medication prediction |
| 2020 | Wang et al.[111] | | | | ✔ | FReaConv | CNN | Heart failure and mortality prediction |
| 2020 | Xiang et al.[112] | | | | ✔ | TSANN | LSTM | Asthma exacerbation prediction |
| 2020 | Yin et al.[113] | | | ✔ | | TAME | LSTM | Sepsis phenotyping |
| 2020 | Yu et al.[114] | | | | ✔ | | LSTM | Mortality prediction |
| 2020 | Yu et al.[115] | | | | ✔ | | LSTM, GRU | Mortality prediction |
| 2020 | Zeng et al.[116] | ✔ | | ✔ | ✔ | MSAM | Encoder | Disease and medical cost prediction |
| 2020 | Zhang et al.[117] | | ✔ | | | HAP | RNN | Procedure and diagnosis prediction |
| 2020 | Zheng et al.[118] | | | | ✔ | Tracer | RNN | AKI and mortality prediction |
| 2020 | Park et al.[16] | | | | ✔ | | RNN, ANN | Bacteremia prediction |
| 2020 | Thorsen-Meyer et al.[119] | | | | ✔ | | LSTM | Mortality prediction |

## 3.2 Challenges and deep learning solutions

The following section will summarize the four major challenges (i.e., data irregularity, data sparsity, data heterogeneity, and model opacity) posed by the temporal EHR data and examine their corresponding deep learning solutions.

### 3.2.1 Data irregularity in temporal EHR

Irregular data is pervasive in temporal EHR [39, 120], where the time intervals between various encounters vary, resulting in challenges modeling the whole time series. These irregular time intervals may contain valuable hidden information. For example, shorter time intervals may imply more frequent examinations, indicating the worsening condition of a patient. To utilize the latent information, researchers usually extract a series of time intervals [3, 46], represented as follows for the $i$-th patient.

$$\boldsymbol{\varphi_{i,(a-1,a)}} = |t_a - t_{a-1}|, a = 1, \cdots, T \qquad (2)$$

Deep learning could naturally capture this long-term sequential effect, and two groups of deep learning solutions have been proposed. One group of methods directly model time series, taking irregular time intervals as the input variables, where customized neural network architectures ingeniously fuse each irregular time point [31, 38, 78, 82]. Although the time lapse between successive elements in patients' records may vary from days to years, novel deep learning approaches can fit the unequally distributed data with their inherent temporal structure (e.g., gate architecture of the LSTM) based on the time distribution and its interval $\boldsymbol{\varphi_{i,(a-1,a)}}$ directly, such as T-LSTM [78]. Moreover, a variety of integrated data processing systems have been proposed based on diverse deep learning structures. For example, Deep Patient [31] utilized a three-layer stack of denoising autoencoders to capture hierarchical regularities and dependency in the temporal coding data. REverse Time AttentIoN (RETAIN) [33] was developed based on a two-level neural attention model to detect influential past visits. Deepr [38] ingeniously transformed a record into a sequence of discrete elements separated by coded time gaps and hospital transfers. Xiang et al. [82] applied dynamic input windows to acquire time-sensitive coding information.

Another group of approaches attempts to transform irregular data into regular ones by determining a fixed interval and then treating the time points without data as missing [121]. While the irregular data could naturally be transformed into regular data with the same time intervals, this strategy may result in many missing values [60, 71, 122] since there are many time intervals without measurements, necessitating imputation. In this situation, a masking vector $u \in \{0, 1\}$ [10] is usually used to represent its missing status. The missing values would reduce statistical power and cause bias in estimating mass parameters in deep learning methods [123].

Most researchers used carry-forward imputation to address the missing value issue in temporal EHR [124], where the last observed values were used for all subsequent missing observation points. However, this solution is likely to introduce bias, as it assumes that the value remains unchanged from the last observation. Traditional imputation methods, such as median imputation and multiple imputation, may not capture hidden patterns in temporal data effectively, calling for deep learning approaches with temporal representations. Recently, Macias et al. [75] proposed a novel imputation method in an ICU setting by exploiting temporal dependencies through autoencoder-represented information. Zhang et al. [85] imputed missing values of multivariate time series by a denoising autoencoder. Based on GRU, Che and colleagues [35] designed GRU-D to utilize informative missingness with prior-based regularization. Furthermore, Jun and colleagues [68] developed a general framework to incorporate effective missing data imputation with a variational autoencoder.

### 3.2.2 Data sparsity in temporal EHR

There are a wide range of concepts in medicine, such as medical diagnosis, medication, and treatment information. We denote them as $(c_{i,t})$, which are the most commonly investigated temporal variables (used in 78 out of 98 included papers). Most medical concepts are mapped to a corresponding coding system, such as the International Classification of Diseases (ICD) [125], Current Procedural Terminology (CPT) [126], and Medication Reference Terminology (MED-RT) [127]. Their sparsity and high dimensionality, however, may hinder analysis. Traditionally, medical ontologies such as SNOMED [128], Charlson Comorbidity Index [129], RxNorm [130], and Logical Observation Identifiers Names and Codes (LOINC) [131] provide structured hierarchical approaches to convert the medical concepts into lower-dimensional representations. Despite this, they are unable to extract abundant relations inherent in the temporal data [36, 76, 132]. As a result, the growth in temporal EHR data raises an unmet need for effective representations of temporal data with structured medical codes.

One-hot encoding [133] is the most straightforward technique, converting these categorical variables into several binary columns, where 'one' indicates the presence of the medical code. Nevertheless, it is not an ideal method for encoding high-dimensional categorical variables. Moreover, one-hot encoding no longer holds the promise of preserving hidden relationships in sparse coding. As an example, pneumonia and bronchitis are highly concurrent, but such an internal relationship between their codes is not reflected in one-hot coding.

Therefore, medical concept embedding has become a mainstream method [32] for learning latent representations from high-dimensional sparse medical codes. We denote a set of medical codes as $C = [c_1, c_2, c_3, \ldots c_n]$ in an EHR dataset and $n$ is the total number of codes within the data, which is usually a large number. A visit $\boldsymbol{v_t}$ comprising of several medical codes can be represented as:

$$\boldsymbol{v_t} \in \{0,1\}^n \qquad (3)$$

Then, the embedding function $f_C$ can be represented as:

$$f_C \colon C \to \mathbb{R}^m \qquad (4)$$

where $m$ represents the dimensionality of the embedding vector and is often a much smaller number than $n$.

Medical code embedding originated from Word2Vec [134], an unsupervised feature extraction method for natural language processing (NLP), which converts words to numerical embedding by mapping each word token into a high-dimensional vector space. The basic form of the one-layer embedding neural network is shown below in equation (5),

$$u_t = ReLU(\boldsymbol{W_c} \boldsymbol{v_t} + \boldsymbol{b_c}) \qquad (5)$$

where $u_t$ is the low-dimensional vector for subsequent downstream tasks, and $\boldsymbol{W_c}$ denotes the embedding matrix. Recently, Choi et al. [135, 136] proposed to learn distributed representations of sparse medical codes (e.g., diagnoses, medications, and procedure codes) using Word2Vec and applied them to several clinical prediction tasks. Subsequently, Med2Vec [137] was proposed to extend the original Word2Vec with a multilayer perceptron for learning both succinct codes and visit-level representations.

Med2Vec was further extended to integrate embedding systems with different deep learning architectures for sparse temporal EHR data. Lu and colleagues [138] proposed utilizing hyperbolic embeddings of medical concepts instead of traditional Euclidean space geometry. Moreover, the cross-field categorical attributes embedding (CCAE) [102] was developed to learn a vectorized representation for cancer patients at attribute-level by orders, where strong semantic coupling among categorical variables was exploited effectively. Esteban et al. [27] employed the Markov model to learn personalized Markov embeddings. MC2Vec [132] was then designed to capture the proximity relationships between medical concepts through a two-step optimization framework that recursively refines the embedding for superior output. Patient2vec [80] introduced the use of the RNN model to learn sequential context-aware features of visits and the correlations between physical symptoms and associated treatments. Zhang et al. [117] developed hierarchical attention propagation (HAP), a hierarchically propagating attention across the entire ontology structure, where a medical code adaptively learns its embedding from all other codes in the hierarchy instead of only its ancestors.

### 3.2.3 Data heterogeneity in temporal EHR

Heterogeneity is another challenge undermining the quality of data analyses, where EHRs generally consist of multiple data modalities and outcomes. In this review, we focused on two types of heterogeneity: patient phenotypes and clinical outcomes.

Patient phenotyping identifies patient sub-cohorts that satisfy complex criteria [139]. Conventionally, phenotypes are identified based on patient similarity, either through statistical distances like Euclidean distance [140] or by machine learning methods such as k-means [32]. However, when dealing with high-dimensional and multimodal longitudinal data, these traditional methods cannot identify complicated patient phenotypes and were unable to retain most long-term temporal information [32]. In contrast, a deep learning structure could capture these complex temporal dynamics in the longitudinal EHR for evaluating patient similarity. The typical deep neural network architectures include CNN [32, 43, 56], LSTM [58], and autoencoder [57, 62]. For instance, Yang et al. [18] utilized the RNN model to improve the phenotyping of asthma using ICU time sequence data.

Another form of heterogeneity is related to the diversity of clinical outcomes and disease conditions, including their complex interrelationships. In this context, a single-task approach (e.g., binary logistic regression) is ineffective. Even though statistics methods, such as multinomial logistic regression, are capable of making a multi-label prediction, deep learning presents great promise for handling them since neurons in neural networks can used for more than one task, aiming to jointly learn multiple prediction tasks simultaneously. As reported in [42] [83] [115], several multitask frameworks were proposed based on RNN, which shared the same networks, but with a task-specific layer to monitor a particular disease or outcome. The advantages include sharing the same layers, saving computing resources, and extracting certain useful temporal information. In addition, Suresh et al. [57] improved the method through a two-step procedure, with the first step being unsupervised clustering through a sequence-to-sequence autoencoder, and the second step being the outcome prediction. Harutyunyan and colleagues [141] summarized the advantages of multitask learning over single-task learning, by proposing clinical prediction benchmarks using temporal EHR data obtained from the MIMIC-III database.

### 3.2.4 Opacity in modeling temporal EHR

While deep learning provides diversified solutions to deal with temporal EHR data, its black box nature presents another significant challenge. Due to the depth of neural network layers and the complexity of each module, understanding sophisticated deep learning models remains elusive, particularly when dealing with temporal data. Many researchers have attempted to explain deep learning models with post hoc explanations (e.g., Doctor XAI) [106, 142], while others advocate that the models themselves should be interpretable [143, 144].

Two groups of approaches have been proposed to interpret black box deep learning models: mimic learning and attention mechanisms. Mimic learning simulates deep learning models through an inherently transparent model such as logistic regression [145]. Gradient boosting trees (GBT) [35] was also used to imitate the process of GRU and achieved superior performance and good interpretability when extracting the importance of features.

The concept of attention mechanism [146], originally derived from NLP and used in machine translation to adjust weights of different words, has been popular in research on temporal EHR [147]. Researchers have used attention to determine which time points in the patients' medical history are more predictive of outcomes [42, 54, 61, 83, 115, 118, 148]. Attention can also provide insight into the importance of different visits or variables for aiding medical decision making, where larger attention represents greater importance. An example from Shi et al. [148] is used here to illustrate the attention mechanism. Given the input data $\boldsymbol{D_i}$ and hidden layer $\boldsymbol{H_i}$ after RNN layers processing, as formulated by equation (6),

$$\boldsymbol{H_i} = [h_{i,1}, h_{i,2}, h_{i,3}, \dots, h_{i,T-1}, h_{i,T}] \qquad (6)$$

the attention $\boldsymbol{\alpha_i}$ is calculated through equation (7), where $\boldsymbol{w}^T$ is the parameter obtained from the model training.

$$\boldsymbol{\alpha_i} = softmax(\boldsymbol{w}^T \tanh(\boldsymbol{H_i})) \qquad (7)$$

Afterward, we obtain the attention tensor $T_i$ for downstream tasks.

$$T_i = \boldsymbol{tan}(\boldsymbol{H_i}\boldsymbol{\alpha_i}^{\mathrm{T}}) \qquad (8)$$

There are two major attention approaches, one of which treats all temporal variables at the same level, while the other takes data hierarchy into account. As shown in [47, 49, 149, 150], neural networks were proposed with the attention layer to calculate the weights in the absence of data hierarchy. Crossover attention model (COAM) [67] was designed through the crossover attention mechanism by leveraging the correlation between diagnosis and treatment information. Park et al. [151] further improved the attention mechanism by adding feature occurrence frequency to capture critical temporal variables that appeared infrequently. Bai et al. [46] enhanced the attention by learning time decay factors, making it possible to interpret the chronic disease progression and understand how the risks of future visits change over time. Furthermore, some models incorporated heterogeneous data such as treatment, medication, procedures, and diagnoses. Specifically, DeepRisk [64] integrated multiple time-ordered clinical data as a whole by handling correlation among predictors via a single DNN and three attention-based LSTMs. KAME [52] was further developed as a knowledge-based attention mechanism, which used medical domain knowledge, and computed the attention based on a directed acyclic graph of various medical concepts.

Another major attention approach exploited the hierarchical structure of temporal data, such as the data from both admission and medical event levels [152]. During the course of admission, a number of events occur, which are usually recorded by medical codes like ICD-9. This hierarchical structure with multiple attention levels was intended to integrate local and global time information and enhance model performance and interpretation. Several studies [33, 40, 105, 112, 116] have applied a two-level neural attention model within RNN, where the first level pertains to medical events, while the second level pertains to visits. A good example is RETAIN [33], which integrates two levels of attention to make use of time information in feature aggregation and explain the critical medical event in the input sequence. To improve the interpretability of RETAIN, Kwon et al. [69] developed RetainVis, an interactive visualization tool. Another attempt was to apply graph-level attention (based on knowledge graph) simultaneously with other attentions [86]. Overall, the hierarchical attention could not only digest the sequence information correctly in temporal EHR, but also provide an insightful interpretation of the importance of each variable or timepoint.

## 4. Discussion

This review summarized the challenges related to temporal EHR data and discussed how deep learning solutions could help to overcome them. While temporal EHR data is valuable for biomedical informatics research, its complex structure poses a challenge to standard learning algorithms. Deep learning models have shown the ability to present temporal data in a novel manner while retaining sequential information efficiently. This study aimed to address a knowledge gap on deep learning for temporal EHR by exploring current challenges and summarizing methodologies. Through a systematic literature review, we identified four major challenges, including data irregularity, sparsity, heterogeneity, and model opacity. During the last decade, nearly one hundred novel deep learning methods have been proposed, and this number continues to grow rapidly over time, demonstrating the importance and potential of deep learning in temporal EHR data analysis. While these deep learning techniques have shown promising results, several challenges remain, including the need for high-quality data and the issue regarding their applicability to clinical practice. Ideally, future studies could consider designing a comprehensive system that combines solutions to all challenges.

Despite various attempts to address data irregularity, heterogeneity and sparsity, there is still a great need for improvements to the data itself. For deep learning algorithms to be successful, large-scale EHR datasets are always required. The most commonly used dataset in our included papers was MIMIC (n=36/98) [25], a well-organized and freely accessible critical care database developed at the Beth Israel Deaconess Medical Center. The majority of studies analyzed only one dataset, and only ten utilized two or more datasets, raising questions about the transferability and generalizability of the models. Therefore, we recommend the development of more large-scale EHR databases that are freely accessible worldwide, providing the opportunity for multicenter validation of current models. Aside from data size, the quality of a dataset is another important factor affecting model performance; improvements in data collection and processing, such as the correction of outliers due to mistyping or misalignment, may be considered.

Although the availability of large, labeled data is always desirable, situations with limited data are common in medical settings because of the costs of labeling and the sensitive nature of their sharing. Many approaches have been proposed to solve this issue, including data augmentation and optimal utilization of data. Generative Adversarial Networks (GANs) provide a compelling solution to amplify temporal data, and Che et al. [34] demonstrated that the newly generated data are of appropriate quality. Zhang et al. [88], on the other hand, managed to take advantage of limited data through the novel representation framework MetaPred based on longitudinal patient EHRs. Another potential solution would be transfer learning, which allows us to transfer knowledge between multiple hospitals or EHRs, and combine various sources to extract knowledge, referred to as multi-source transfer [153]. General transfer learning consists of a two-stage paradigm [154], where the leading deep learning network is generally trained on a large-scale, publicly available benchmark dataset. Next, the pre-trained network is further conditioned on the specific local data with limited samples. Transfer learning has the potential to relieve the data shortage in healthcare and improve the model's generalizability [155].

There has been considerable discussion of the opacity issue for temporal deep learning models, especially for medical applications where there are high-stakes decisions. Several attention mechanisms have been proposed to address these black box models [33, 40, 105, 112, 116]. Grad-CAM [156] is a widely adopted algorithm developed initially to provide visual explanations for CNN by highlighting the important regions, and was recently extended to the medical field [157]. However, these post-hoc interpretability approaches may lead to explanations resulting from certain artifacts learned by the model rather than actual knowledge derived from the data [158]. This limitation raises concerns on model usability in actual healthcare settings. As a comparison, ante-hoc interpretable models are preferred by doctors and nurses in clinical practice because they can understand them naturally and inherently [143, 159, 160]. Recently, Ustun and Rudin developed the Risk-calibrated Supersparse Linear Integer Model (RiskSLIM) [161] and further improved it through the optimization of risk scores [162]. Apart from that, Xie et al. provided practical solutions, AutoScore [163] and its extensions [164, 165], by leveraging interpretable machine learning for clinical score generation. These intrinsically interpretable methods have considerable potential for integrating deep learning techniques to facilitate model validation in real-world settings.

In addition to CNN (n=9), RNN (n=18), and LSTM (n=35) as the most commonly used deep learning architectures, Transformer [166] and MLP-Mixer [167] have recently emerged as popular alternative frameworks. Transformer computes temporal representations entirely on the basis of self-attention without the use of sequence-based RNNs or convolution. It has shown great potential in temporal EHR representation [101, 105]. Furthermore, MLP-Mixer [167] was later developed with a simpler structure, not requiring any convolutions or attention. With one MLP for per-location features and another for spatial information, MLP-Mixer appears to be a conceptually and technically succinct alternative for processing temporal EHR data.

This study has several limitations. First, we sought to understand the current state of the literature from a methodological perspective. Still, we did not attempt to summarize all clinical applications and report the performance of deep learning solutions. Second, considering the heterogeneity of data pre-processing, parameter tuning approaches, and clinical tasks among the included studies, we were unable to recommend the overall best deep learning methods for temporal EHR data analysis. Third, this review focused exclusively on deep learning methods for analyzing structured temporal EHR data. It will be beneficial in the future to investigate techniques that deal with both structured and unstructured data (e.g., clinical notes, medical images, and physiological signals). Lastly, the exclusion of preprints in our analysis may have overlooked some new evidence but was able to ensure the inclusion of only peer-reviewed scientific results.

## 5. Conclusion

We comprehensively reviewed the primary issues in analyzing temporal EHR data and presented state-of-the-art deep learning solutions. Various significant challenges arising from the representation of EHR temporal data were addressed to some extent by current solutions. Future research may focus on model transferability, clinical domain knowledge incorporation into study design, and model interpretability enhancement to facilitate clinical implementation.

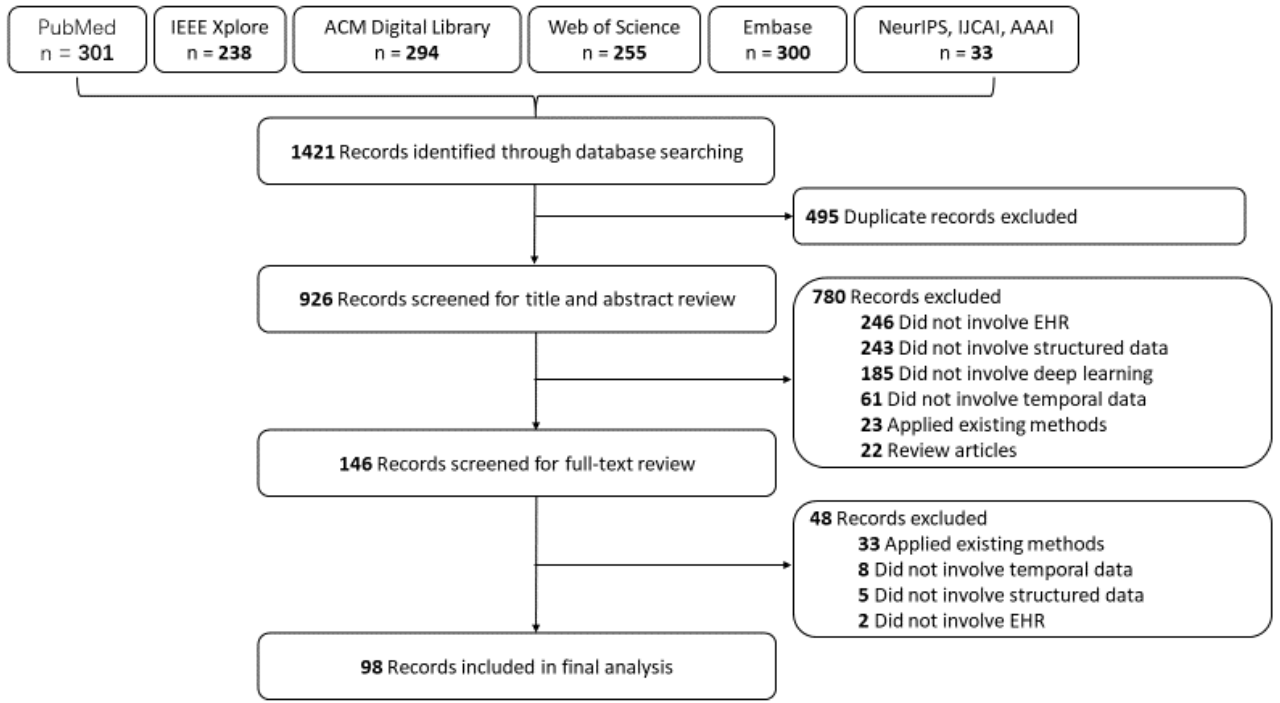

Figure 1: Literature selection flow of deep learning models in temporal EHR data

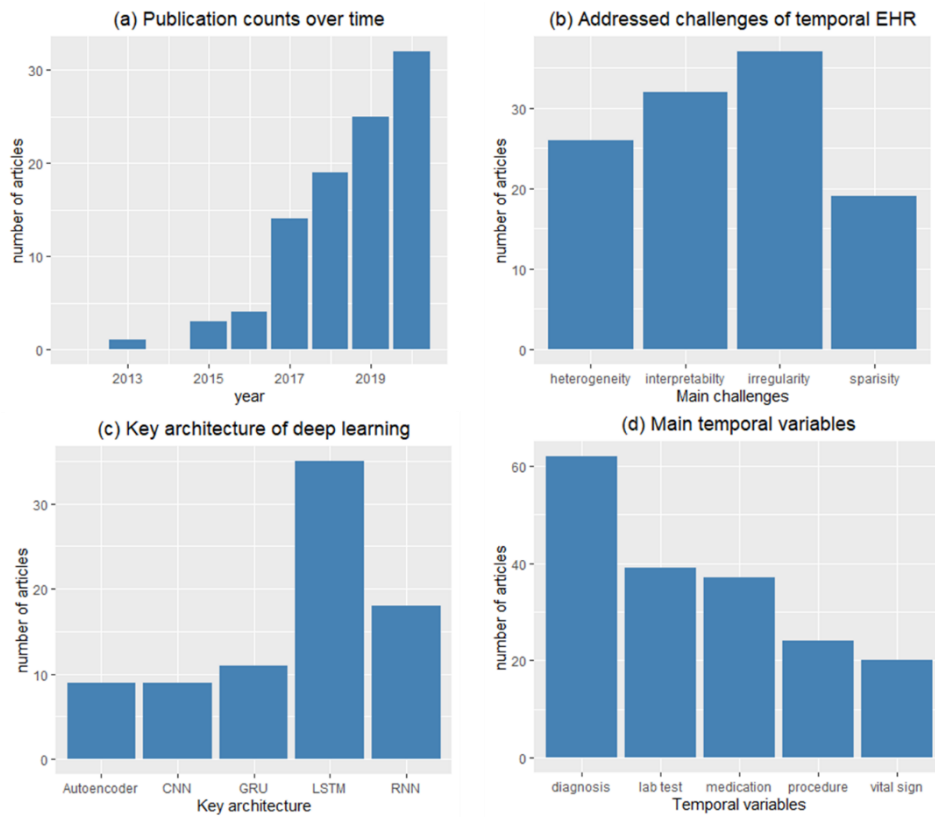

Figure 2: Summarized statistics for all included papers

## Abbreviation:

ANN Artificial Neural Network
AUC Area Under Curve
Bi-LSTM Bi-directional Long Short-Term Memory
Bi-RNN Bi-directional Recurrent Neural Network
BOW Bag of Words
BPR Bayesian Personalized Ranking
CEL Cross-entropy Loss
cFSGL Least Convex Fused Group Lasso
CNN Convolutional Neural Network
DAE Denoising Autoencoder
DDI Drug-Drug Interaction
DPCA Binary Coding with Dimensionality Reduction
DT Decision Tree
EM Expectation Maximization
F1 F1 score
F2 F2 score
FFL Feed-Forward Layer
FNR False Negative Rate
FPR False Positive Rate
GAN Generative Adversarial Network
GBDT Gradient Boosting Decision Tree
GBM Gradient Boosting Machine
GBRT Gradient Boosting Regression Tree
GCN Graph Convolutional Network
GMM Gaussian Mixture Model
GNN Graph Neural Network
GRAM Graph-based Attention Model
GRU Gated Recurrent Unit
HP Hawkes Process
KNN K-nearest neighbors method
LDA Latent Dirichlet Allocation
LM Linear Model
LR Logistic Regression
LSTM Long Short-Term Memory
MAE Mean Absolute Error
MCC Matthews Correlation Coefficient
MEWS Modified Early Warning Score
MF MissForest
MLP Multi-Layer Perceptron
Mod Modularity
MSE Mean Squared Error
NBN Naive Bayesian Network

nFSGL Least Non-Convex Fused Group Lasso
NMI Normalized Mutual Information
PCA Principal Component Analysis
PPV Positive Predictive Value
PRC Precision Recall curve
RBM Restricted Boltzmann Machine
RF Random Forest
RMSE Root Mean Squared Error
RNN Recurrent Neural Network
ROC Receiver Operating Characteristic Curve
SDAE-SM SDAE appending a softmax layer
Sensitivity = Recall
SIRS systemic inflammatory response syndrome
SOFA Sequential Organ Failure Assessment
Spec Spectral Clustering
SVM Support Vector Machine
TANN Time-aware neural network
TCN Temporal Convolutional Network
TNR True Negative Rate
TPR True Positive Rate

# Appendix

eTextbox: Search strategy for each database

## PubMed

('deep learning' OR 'neural network' OR 'deep' OR 'CNN' OR 'RNN' OR 'LSTM') AND ('embed' OR 'embedding' OR 'representation' OR 'time series' OR 'sparse' OR 'temporal' OR 'concept' OR 'sequential' OR 'attention') AND ('electronic health record' OR 'EHR' OR 'EHRs' OR 'electronic medical record' OR 'EMR' OR 'EMRs')

## IEEE Xplore

('deep learning' OR 'neural network' OR 'deep' OR 'CNN' OR 'RNN' OR 'LSTM') AND ('embed*' OR 'representation' OR 'time series' OR 'sparse' OR 'temporal' OR 'concept' OR 'sequential' OR 'attention') AND ('electronic health record*' OR 'EHR*' OR 'electronic medical record*' OR 'EMR*')

## ACM Digital Library

("deep learning" OR "neural network" OR "deep" OR "CNN" OR "RNN" OR "LSTM") AND ("embed*" OR "representation" OR "time series" OR "sparse" OR "temporal" OR "concept" OR "sequential" OR "attention") AND ("electronic health record*" OR "EHR*" OR "electronic medical record*" OR "EMR*")

## Web of Science

((TS = (deep learning OR neural network OR deep OR CNN OR RNN OR LSTM)) AND (TS = (embed* OR representation OR time series OR sparse OR temporal OR concept OR sequential OR attention)) AND (TS = (electronic health record* OR EHR* OR electronic medical record* OR EMR*)))

## Embase

("deep learning" OR "neural network" OR "deep" OR "CNN" OR "RNN" OR "LSTM") AND ("embed*" OR "representation" OR "time series" OR "sparse" OR "temporal" OR "concept" OR "sequential" OR "attention") AND ("electronic health record*" OR "EHR*" OR "electronic medical record*" OR "EMR*")